\documentclass{article} 
\usepackage{amsmath}
\usepackage{iclr2025_conference,times}
\usepackage{multirow}

\usepackage[table]{xcolor}
\usepackage[justification=raggedright,singlelinecheck=false]{caption}

\usepackage[utf8]{inputenc} 
\usepackage[T1]{fontenc}    
\usepackage{newunicodechar}
\newunicodechar{∼}{\textasciitilde}
\usepackage{hyperref}       
\usepackage{url}            
\usepackage{booktabs}       
\usepackage{amsfonts}       
\usepackage{nicefrac}       
\usepackage{microtype}      
\usepackage{graphicx}       
\usepackage{xcolor}         
\usepackage{amsmath}   
\usepackage{amssymb}   
\usepackage{fontawesome} 
\usepackage{listings}
\usepackage{listingsutf8}
\usepackage{wrapfig}
\usepackage{ragged2e} 

\usepackage{xcolor}
\definecolor{good}{HTML}{1B5E20}   
\definecolor{bad}{HTML}{C62828}    
\newcommand{\good}[1]{\textcolor{good}{\textbf{#1}}}
\newcommand{\bad}[1]{\textcolor{bad}{\textbf{#1}}}

\usepackage[most]{tcolorbox}
\tcbuselibrary{skins,breakable,listingsutf8}

\usepackage{amsmath,amsfonts,bm}

\def\eqref#1{equation~\ref{#1}}

\def\1{\bm{1}}

\DeclareMathAlphabet{\mathsfit}{\encodingdefault}{\sfdefault}{m}{sl}
\SetMathAlphabet{\mathsfit}{bold}{\encodingdefault}{\sfdefault}{bx}{n}

\usepackage{hyperref}
\usepackage{url}

\title{\raggedright Fathom-DeepResearch: Unlocking Long \mbox{Horizon} Information Retrieval and Synthesis for SLMs}

\author{%
  Shreyas Singh* \\
  Fractal AI Research \\
  \texttt{shreyas.singh@fractal.ai} \\
  \And
  Kunal Singh{*\dag} \\
  Fractal AI Research \\
  \texttt{kunal.singh@fractal.ai} \\
  \And
  Pradeep Moturi* \\
  Fractal AI Research \\
  \texttt{pradeep.moturi@fractal.ai} \\
}

\iclrfinalcopy 
\begin{document}

\maketitle

\begin{abstract}

Tool-integrated reasoning has emerged as a key focus for enabling agentic applications. Among these, DeepResearch Agents have gained significant attention for their strong performance on complex, open-ended information-seeking tasks. We introduce Fathom-DeepResearch, an agentic system composed of two specialized models. The first is Fathom-Search-4B, a DeepSearch model trained from Qwen3-4B and optimized for evidence-based investigation through live web search and targeted webpage querying. Its training combines three advances: (i) DUETQA, a ∼5K-sample dataset generated via multi-agent self-play that enforces strict web-search dependence and heterogeneous source grounding; (ii) RAPO, a zero-overhead extension of GRPO that stabilizes multi-turn Reinforcement Learning with Verifiable Rewards through curriculum pruning, reward-aware advantage scaling, and per-prompt replay buffers; and (iii) a steerable step-level reward that classifies each tool call by cognitive behavior and marginal utility, enabling explicit control over search trajectory breadth, depth, and horizon. These improvements enable reliable extension of tool-calling beyond 20 calls when warranted. The second is Fathom-Synthesizer-4B, trained from Qwen3-4B, which converts multi-turn DeepSearch traces into structured, citation-dense DeepResearch Reports for comprehensive synthesis. Evaluated on DeepSearch benchmarks (SimpleQA, FRAMES, WebWalker, Seal0, MuSiQue) and DeepResearch-Bench, the system achieves state-of-the-art performance in the open-weights category while demonstrating strong generalization to diverse reasoning tasks including HLE, AIME-25, GPQA-Diamond, and MedQA.
\end{abstract}

\begin{center}
\faGithub \href{https://github.com/FractalAIResearchLabs/Fathom-Search}{https://github.com/FractalAIResearchLabs/Fathom-DeepResearch}
\end{center}

\begingroup
\renewcommand\thefootnote{\fnsymbol{footnote}}
\setcounter{footnote}{1}%
\footnotetext{Equal contribution.}
\stepcounter{footnote}%
\footnotetext{Project lead.}
\endgroup

\section{Introduction}

\begin{figure*}
  \centering
  \includegraphics[width=13cm]{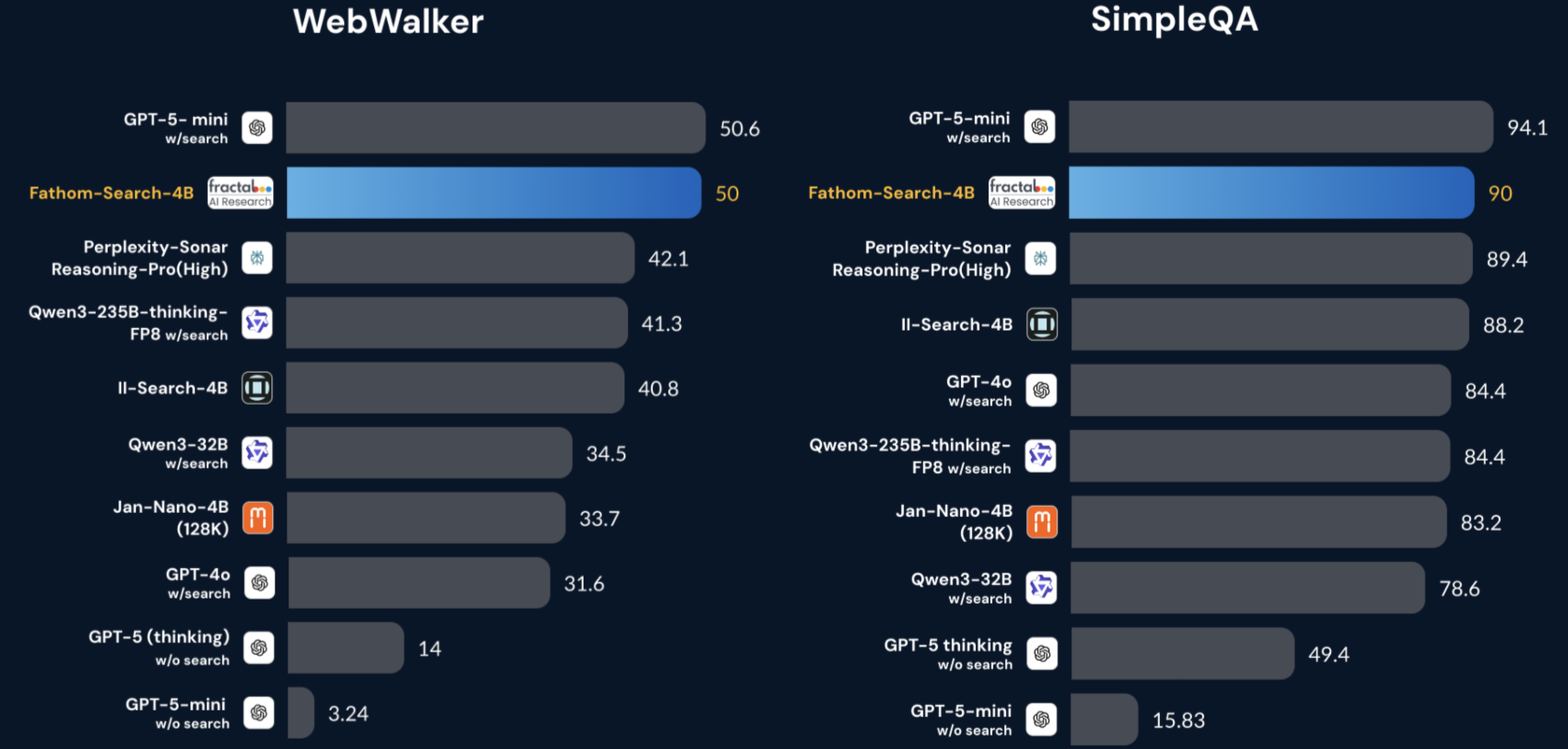}
  \caption{Comparison of \textbf{Fathom-Search-4B}  on prominent DeepSearch benchmarks \textit{(WebWalker, SimpleQA)}. Our model consistently outperforms strong open-source \& closed source baselines.}
\end{figure*}

Recent advancements in reasoning capabilities of Large Language Models (LLMs) have enabled a significant performance advancement across a diverse set of tasks, such as mathematical reasoning, code generation \citep{jain2024livecodebench, deepseekai2025deepseekr1incentivizingreasoningcapability, singh2025sbscstepbystepcodingimproving, singh2025no}. We are not only witnessing expert level performance on academic benchmarks, but are perceiving a paradigm shift towards agentic intelligence. Owing to tool-integrated reasoning, these models can now autonomously observe, reason and interact with complex and dynamic environments. Contemporary state-of-the-art tool-augmented AI systems like DeepResearch \citep{openaideepresearch}, have exhibited super-human performance in highly sophisticated long-horizon, deep-information retrieval and synthesis tasks \citep{deepresearchbench}. These agents transcend the limitations of static parametric knowledge by implementing dynamic reasoning frameworks that autonomously partition multifaceted queries, coordinate multiple tool interactions, and integrate heterogeneous information sources into unified, evidence-supported conclusions.

However, a substantial performance gap remains between proprietary implementations \citep{openaideepresearch, geminideepresearch, perplexitydeepresearch} and open-source \citep{langchainopendeepresearch,kimiresearcher} alternatives, making the development of robust DeepResearch architectures a critical challenge. Current open-source frameworks suffer from two fundamental limitations. First, they lack proficiency in sustained tool usage required for high-uncertainty reasoning and synthesis tasks \citep{deepresearchbench}. Efforts to scale DeepResearch capabilities are constrained by (i) the absence of a high-quality, verifiable, and scalable dataset creation pipeline, (ii) algorithmic instability in multi-turn reinforcement learning (RL) with tools, and (iii) inefficient tool-calling behavior that undermines deep information exploration and retrieval. Second, there is an overemphasis on closed-form problem solving, which comes at the expense of the information synthesis capabilities essential for tackling open-ended investigative queries.
In the subsequent section we discuss the aforementioned issues in detail.

\subsection{Motivation}
\label{Motivation}

\noindent\textbf{(1) Training instability of GRPO in multi-turn tool interaction}: RLVR (Reinforcement Learning with verifiable rewards) with GRPO \citep{DSMgrpo} has demonstrated early promise in aligning LLMs with sparse reward signals for single-turn reasoning tasks, particularly in structured domains like Math/STEM \citet{DSMgrpo, yang2024qwen25mathtechnicalreportmathematical}. However, GRPO struggles to scale to multi-turn tool-augmented environments, because external tool interaction responses induce distribution shift in the policy model from its set token generation patterns, this leads to decoding instability and malformed generations. This cascading of errors causes group-relative advantages to saturate, leading to extremely  unstable gradient updates that breaks the entire training process. \citep{simpletir}.

\noindent\textbf{(2) Reward hacking and inefficient tool calling}
\noindent\textit{(a) Correctness-only sparse rewards do not scale to long-horizon tool calling.} When training with only a single end of episode correctness signal, the agent shows early improvements achieving format adherence and basic tool-calling competence in the beginning, however, as training progresses, tool usage increases sharply while both training reward and validation performance deteriorate \citep{sfrdr}. This degradation stems from reward hacking: the agent collapses into repetitive, identical tool calls because the vanilla RLVR objective provides no incentive for efficiency or diversity in tool use. \noindent\textit{(b) RL amplifies SFT priors, limiting control over the cognitive behaviors developed by the policy} \citep{cog}: Tool-use RL typically relies on an SFT cold start to elicit basic tool competence \citep{li2025websailornavigatingsuperhumanreasoning}; \citep{ARPO} RL then amplifies pre-existing cognitive behaviors seeded by SFT. Standard RLVR affords limited control over the exploration and verification strategies developed by the policy model, consequently the quality of cold-start trajectories disproportionately shape the policy model's tool-use behavior and provides no steerability.

\noindent\textbf{(3) Limited training data characterized by high and hard-to-reduce intrinsic information uncertainty}: Training datasets such as TriviaQA \citep{2017arXivtriviaqa}, and multi-hop variants like \text{2WIKI}\citep{xanh2020_2wikimultihop}, and HotpotQA \citep{yang2018hotpotqa} represent problems where solutions can often be found through minimal set queries or even from a model’s parametric knowledge alone. These datasets do not expose models to the real-world retrieval challenges posed by noisy, heterogeneous data sources on the internet. Recent synthetic efforts \citep{sun2025zerosearchincentivizesearchcapability, li2025websailornavigatingsuperhumanreasoning, sun2025simpledeepsearcher} attempt to bridge this gap by simulating realistic search behavior. For instance, WebSailor’s\citep{li2025websailornavigatingsuperhumanreasoning} SailorFog-QA constructs ambiguous queries using obfuscated subgraphs of entity graphs, while SimpleDeepResearcher \citep{sun2025simpledeepsearcher} issues multi-stage search-summarize-generate tool calls over raw HTML. Despite their innovation, these pipelines remain expensive, brittle, and time-consuming. They rely on handcrafted heuristics, graph expansion, or multi-stage LLM orchestration, limiting scalability, topical diversity, and adaptability to new domains. 

\noindent\textbf{(4) Challenges in handling open-ended qureies}
Recent efforts \citep{dao2025jannanotechnicalreport,II-Search-4B,li2025websailornavigatingsuperhumanreasoning} primarily focus on optimizing model performance for closed-ended queries, which are characterized by well-defined objectives. However, many real-world applications demand handling of open-ended, exploratory queries that fundamentally differ from their closed-ended counterparts. These questions lack singular definitive answers and  hence, not only demand extensive multi-turn exploration to uncover diverse perspectives, but also require synthesis of comprehensive responses that integrate multiple findings with rigorous evidence grounding. The inherently exploratory nature of these problems necessitates sophisticated information synthesis capabilities, a critical gap that existing approaches fail to address.

\begin{figure*}[t]
  \centering
  \includegraphics[width=\linewidth]{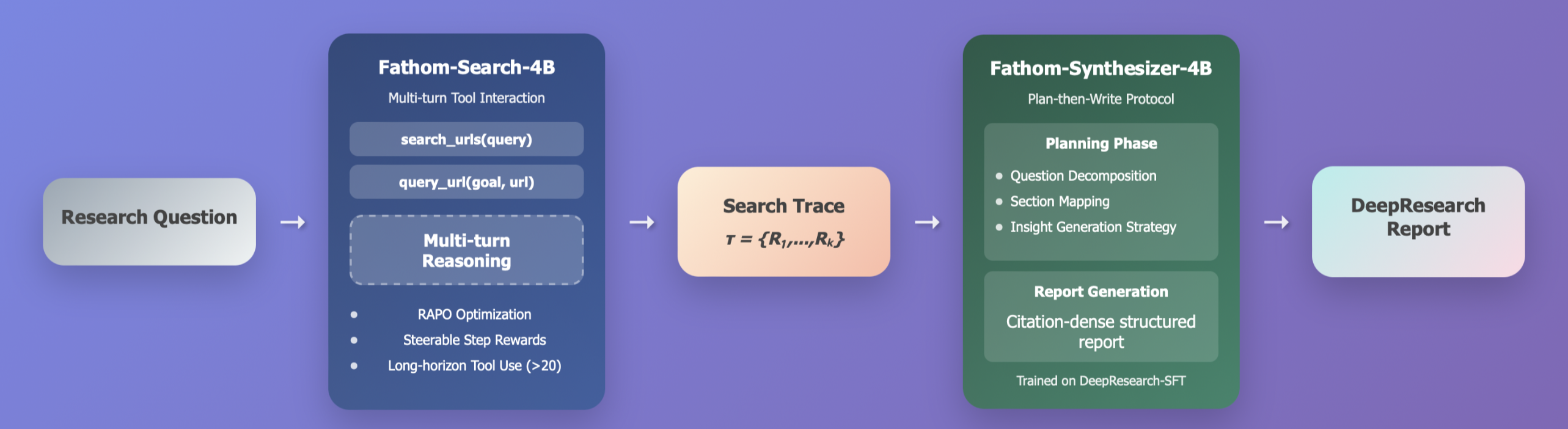}
\caption{\textbf{End-to-end inference framework of our proposed Fathom-DeepResearch} agentic system combining browsing and information gathering ability of \emph{Fathom-Search-4B }with information synthesis and insight generation ability of \emph{Fathom-Synthesizer-4B}}
\label{fig:main}
\vspace{-1em}
\end{figure*}

\subsection{Our Contributions}
To this end, we present an end-to-end DeepSearch system centered on \textit{Fathom-Search-4B} (search enabled reasoning) and \textit{Fathom-Synthesizer-4B} (synthesis \& report-generation). Our key contributions:

\begin{itemize}
\item \textbf{RL Zero framework for DeepSearch training.} We present a novel two-stage RL-Zero framework that helps to \emph{steer cognitive behaviors} developed by the policy model like  exploration \& verification during the training.

\item \textbf{RAPO: Reward Aware Policy Optimization.} We introduce a zero-overhead modification of GRPO with \emph{dataset pruning, advantage scaling, and replay buffers, and a steerable step-level reward} that stabilizes multi-turn RL and enables long-horizon tool use.

\item\textbf{\textsc{DuetQA.}} We release a 5K sample dataset created through our novel \emph{multi-agent self-play pipleline}, which has verifiable question-answer pairs, impossible to answer without \emph{live web search}, for DeepSearch model training.

\item \textbf{\textsc{DeepResearch-SFT.}} A synthetic SFT corpus for converting downstream search/investigation traces of DeepSearch enabled models into DeepResearch reports via an explicit \emph{plan then write} protocol. 

\end{itemize}

\section{Fathom-Search-4B}
\label{Methodology}

We describe the methodology underlying \emph{Fathom-Search-4B}, a tool-using LLM that leverages live web-search capabilities to do evidence based reasoning in a multi-turn tool interaction setting, unlocking long-horizon tool use (\(>\!20\) calls) ability.
These capabilities arise from a combined approach of: (i) a curated synthetic data pipeline tailored to search-tool augmented reasoning, (ii) targeted upgrades to GRPO to effectively adapt it to multi-turn tool interaction, and (iii) a two-stage training regimen with reward shaping to expand the tool-use horizon in a steerable manner.

\subsection{DuetQA: A DeepSearch dataset, generated via multi-agent self play}
\label{sec:dataset-curation}

\begin{figure*}
  \centering
  \includegraphics[width=\linewidth]{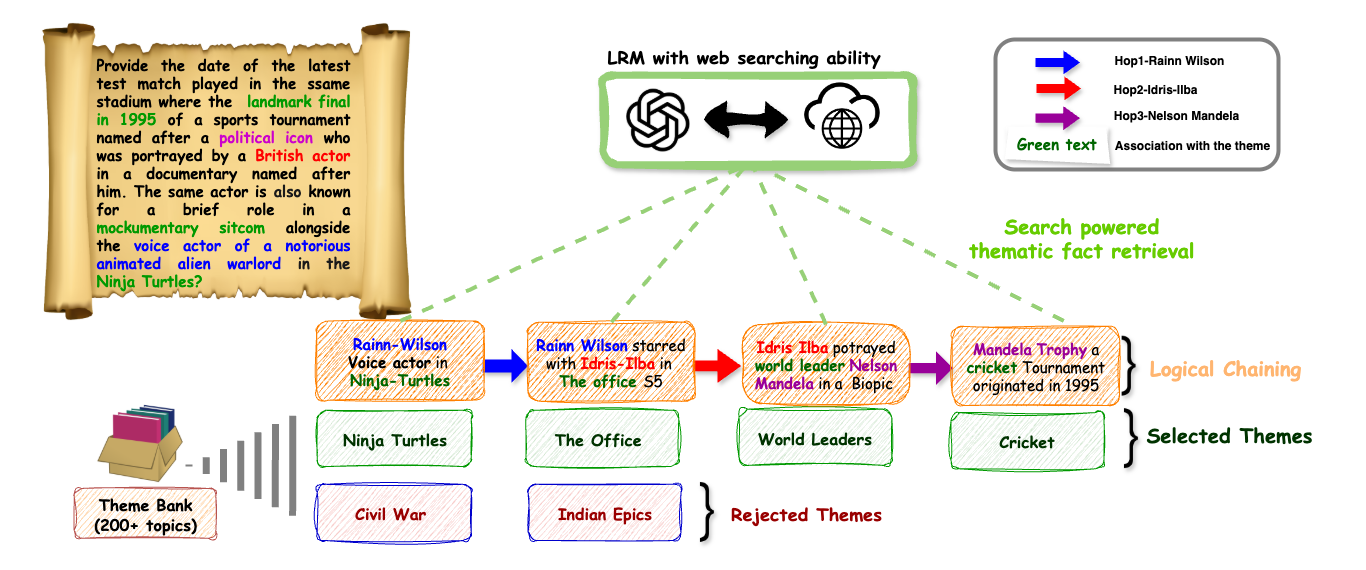}
\caption{\textbf{Multi-agent self-play} framework used to generate a sample  multi-hop DeepSearch question with live-web-search dependency for the \textbf{DuetQA} dataset}
\label{fig:mix}
\end{figure*}

To address the aforementioned dataset challenges in (Sec. \ref{Motivation}), we develop a self-supervised dataset construction framework designed to yield verifiable, search-dependent, multi-hop QA pairs. This pipeline serves as the basis for generating \textsc{DuetQA}, a dataset tailored for training agentic deepsearch models. The design goals are: \textbf{Live web-search dependency}: for each QA pair $(q,a)$, the question is unanswerable without search by enforcing that at least one hop contains information post–2024-01-01 (i.e., for a model $\mathcal{M}$, $P(a \mid q,\mathcal{M}_{\text{no-search}})\ll P(a \mid q,\mathcal{M}_{\text{search}})$); \textbf{Diverse source domains}: questions require querying hetrogeneous web-sources beyond Wikipedia ; and \textbf{Steerable theme control}: each example is grounded in $k\!\in\![5,7]$ sampled themes from  $\mathcal{T}$, a manually curated taxonomy of 200\,+ themes covering a broad range of topics. We generate questions using two frontier web search enabled LRMs, $\mathcal{M}_1$ (O3) and $\mathcal{M}_2$ (O4-mini) \citep{o3_o4mini}, acting as \emph{proxy web-crawling agents} that produce QA pairs and as \emph{independent verifiers} to that ensure question solvability; a third model, $\mathcal{M}_3$ (GPT-4o), is a \emph{non-search} model used for controlled paraphrasing/obfuscation of questions and as a baseline verifier without search.

\noindent\textbf{Data Generation.}  
We adopt two strategies to synthesize multi-hop, search-dependent question-answer pairs. In both, we sample a set of themes 
\(\mathcal{T}_{\text{sample}}\sim\mathrm{Uniform}(\mathcal{T})\) with \(|\mathcal{T}_{\text{sample}}|=k\), \(k\in\{5,6,7\}\).  In the \emph{Mixture of Themes} setting, for each \(t\in\mathcal{T}_{\text{sample}}\), the generator (\(\mathcal{M}_1\) or \(\mathcal{M}_2\)) issues live queries to retrieve recent and/or obscure facts, and composes a multi-hop pair \((q,a)\) by chaining a subset of them into a coherent reasoning path. In the \emph{Seeded Question} setting, we maintain a seed bank of 100 questions; given a seed \(q_0\), the generator rewrites it into a new question \(q\) by integrating one or more sampled facts while preserving the multi-hop scaffold of \(q_0\). In both settings, we enforce that at least one incorporated fact references information after 2024.


\noindent\textbf{Data obfuscation.}
To remove surface cues that let models \emph{short-circuit} the intended multi-hop reasoning, we apply a dedicated obfuscation pass after question generation. Using the non-search model $\mathcal{M}_3$ (GPT-4o) under an in-context learning setup with exemplars, we paraphrase the question to mask intermediate hops. Concretely, $\mathcal{M}_3$ softens exact anchors in each hop by (i) converting specific dates to coarse intervals (``March 2025'' $\rightarrow$ ``early 2025''), (ii) mapping precise numerics to qualitative magnitudes (``1\%'' $\rightarrow$ ``negligible''), (iii) replacing named entities with indirect descriptors (``University of Florida'' $\rightarrow$ ``a major southeastern university''), and (iv) embedding causal/comparative pivots as descriptors rather than explicit connectors. These edits suppress shortcut signals without altering the underlying facts that must be recovered via search.

\noindent\textbf{Multi-agent Verification.} 
We retain a candidate pair $(q,a)$ only if two independent search-enabled LRMs $\mathcal{M}_1$ and $\mathcal{M}_2$ produce the same correct answer while a strong non-search baseline $\mathcal{M}_3$ fails.
This filter enforces correctness through cross-model agreement and certifies that web retrieval is indispensable, ensuring the non-triviality of the question while guarding against overlap of information with the model's parametric knowledge.

\begin{figure*}
  \centering
  \includegraphics[width=\linewidth]{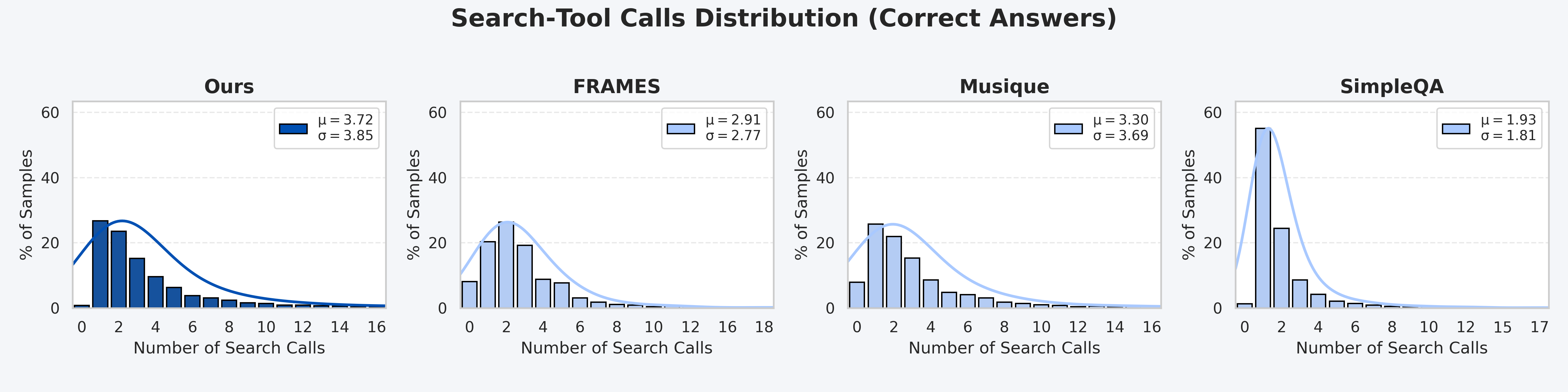}
\caption{Distribution of number of search-calls issued by o3\citep{o3_o4mini} over correctly answered questions comparing DuetQA to other prominent benchmarks. \textbf{DuetQA shows strict live-web-search dependence and multi-hop reasoning} as evident from the long-trained distribution (unlike simpleQA \citep{simpleqa}) and $\ge$ 1 search call(s) required to answer all DuetQA questions correctly (unlike FRAMES \citep{frames}, Musique\citep{musique}).}
\label{fig:fml3}
\end{figure*}

\subsection{Agentic Reinforcement Learning}
\label{subsec:agentic-rl}

In this section, we formulate multi-turn, tool-augmented RL with LLM policies. Let $x\!\in\!\mathcal{X}$ be an input from distribution $\mathcal{D}$ and $\mathcal{T}$ the set of available tools. The policy $\pi_\theta$ generates a reasoning trajectory $\mathcal{R}$ interleaved with tool feedback, followed by a final textual answer $y$. A reference policy $\pi_{\mathrm{ref}}$ is used for KL regularization, and a verifiable reward function $r_\phi$ (LLM-as-judge) provides supervision. The joint rollout can be written as:
\begin{equation}
\label{eq:rollout-factorization}
P_\theta(\mathcal{R}, y \mid x;\mathcal{T})
=
\Big[\prod_{t=1}^{t_{\mathcal{R}}}
P_\theta(\mathcal{R}_t \mid \mathcal{R}_{<t},x;\mathcal{T})\Big]
\cdot
\Big[\prod_{t=1}^{t_y}
P_\theta(y_t \mid y_{<t},\mathcal{R},x;\mathcal{T})\Big],
\quad
\mathcal{R}_t=(\varphi_t,c_t,o_t),
\end{equation}
where $\varphi_t$ is a latent ``\texttt{think}'' segment, $c_t\in\mathcal{T}$ a tool call (with arguments), and $o_t$ the tool response, all expressed in a ReAct-style template. We optimize the policy model with a token-level clipped loss defined as follows:
\begin{equation}
\label{eq:grpo}
\mathcal{L}_{\mathrm{GRPO}}
=
\frac{1}{G}\sum_{i=1}^{G}\frac{1}{T_i}\sum_{t=1}^{T_i}
\min\!\Big[
r_{i,t}\,\hat{A}_{i,t},\;
\operatorname{clip}\!\big(r_{i,t},1-\epsilon,1+\epsilon\big)\,\hat{A}_{i,t}
\Big],
r_{i,t}=\frac{\pi_\theta(o_{i,t}\mid x,\mathcal{H}_{t-1})}{\pi_{\theta_{\text{old}}}(o_{i,t}\mid x,\mathcal{H}_{t-1})}
\end{equation}

For a group of $G$ sampled rollouts with scalar rewards $\{r_i\}$, group-relative advantages  defined as:
\begin{equation}
\label{eq:grpo-adv}
\hat{A}_{i,t}=\frac{r_i-\mu_R}{\sigma_R},\quad
\mu_R=\tfrac{1}{G}\sum_{j=1}^G r_j,\quad
\sigma_R=\sqrt{\tfrac{1}{G}\sum_{j=1}^G(r_j-\mu_R)^2}.
\end{equation}
The trajectory-level scalar reward combines a format score and an answer score:
\begin{equation}
\label{eq:vanilla_reward}
r_i=0.1*R^{\mathrm{format}}_i+0.9*\,R^{\mathrm{answer}}_i
\end{equation}
Here, \(R^{\mathrm{format}}_i\) verifies that rollout follows the  ReAct template (i.e., all steps are correctly wrapped in \texttt{<think>}, \texttt{<tool\_call>}, \texttt{<tool\_response>} tags).  
Meanwhile, \(R^{\mathrm{answer}}_i = \mathbf{1}[a^{(i)}_{\mathrm{pred}} = a_{\mathrm{gt}}]\), where correctness of the final answer is judged by an LLM-as-judge against the ground truth.

\subsection{Agentic Tool Design}
We provide our policy model access to two tools:

\noindent\textbf{\texttt{search\_urls} (web search).}
The tool takes as input a natural language query $q$ and returns a ranked list of triples $(u,\texttt{title},\texttt{snippet})$ using a live search engine. The policy model uses this to identify promising sources and optionally select a URL $u$ for opening in the next step. The tool is invoked as follows: \verb|<tool_call>{name: search_urls, args: {query: q}}</tool_call>|

\noindent\textbf{\texttt{query\_url} (goal-conditioned page reading).}
Given a goal $g$ and a URL $u$, the tool leverages a query LLM to return targeted evidence-backed response that address $g$. This tool enables precise grounding of facts and targeted querying of web-pages. Compared to the injection of entire web-page into the policy model's trajectory, this tool minimizes noise and increases recall. The tool is invoked as follows: \verb|<tool_call>{name: query_url, args: {goal: g, url: u}}</tool_call>|

\subsection{RAPO: Reward-Aware Policy Optimization}
\label{sec:training:failures}

\begin{figure*}
  \centering
  \includegraphics[width=\linewidth]{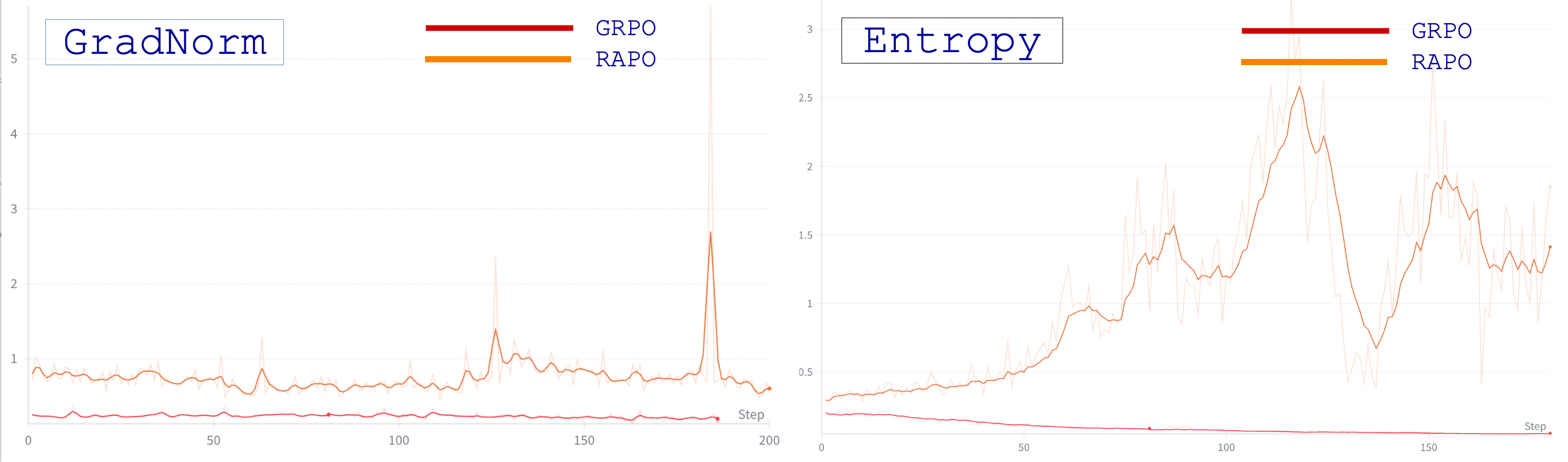}
\caption{Comparison of policy entropy and gradient norm during RLVR training. GRPO exhibits
rapid entropy collapse and diminished gradient norms due to sparse rewards, whereas RAPO sustains
exploration and stronger updates via targeted updates}
\label{fig:gradnorm}
\end{figure*}

In GRPO, the per-prompt (group) reward variance \(\sigma_R\) (Eq.~\ref{eq:grpo-adv}) determines the strength of the advantage signal. When \(\sigma_R{=}0\), group-relative advantages vanish, collapsing batch gradient norms and destabilizing updates (Fig. \ref{fig:gradnorm}). Such \emph{Bad} groups arise under both prompt saturation (all rollouts succeed) and cascading errors (all fail). we inroduce \textbf{RAPO} a lightweight, zero additional rollout cost extension of GRPO designed to stabilize multi-turn, tool-augmented training by mitigating \emph{Bad} group \(\sigma_R{=}0\) pathologies while also preserving sample efficiency.

\textbf{Dataset pruning.}  
First, we prune prompts that are effectively solved at the end of each epoch (Eq. \ref{eq:rapo-pruning}). This prevents training batches from being dominated by saturated groups that provide negligible variance, while implicitly yielding a curriculum in which the active set concentrates on harder prompts.
\begin{equation}
\label{eq:rapo-pruning}
\mathrm{SolveRate}(q)=\tfrac{1}{G}\sum_{i=1}^G \mathbf{1}[R_i>0],\qquad
\mathrm{prune}(q)\Leftrightarrow \mathrm{SolveRate}(q)\ge0.9
\end{equation}
\textbf{Advantage scaling.}  
Second, to counter the dilution of gradients when only a few groups in a batch are informative, we rescale token-level advantages of Good groups inversely with their batch frequency (Eq. \ref{eq:rapo-scale})
This adjustment preserves effective gradient magnitude without requiring costly re-sampling as in DAPO~\citep{dapo}, ensuring that updates remain stable even when informative groups are sparse.
\begin{equation}
\label{eq:rapo-scale}
\tilde{A}_{i,t}=\tfrac{G}{G_{\text{good}}}\,\hat{A}_{i,t},
\qquad
G_{\text{good}}=\#\{\text{groups with }\sigma_R>0\}.
\end{equation}

\textbf{Replay buffer.}  
Finally, we maintain a per-prompt buffer \(\mathcal{B}\) containing the most recent successful trajectory \(\mathbf{o}^\star\) with \(R(q,\mathbf{o}^\star){>}0.5\). If all rollouts for a prompt fail in the current epoch, one trajectory is randomly replaced with \(\mathbf{o}^\star\) from \(\mathcal{B}\). This reintroduces variance (\(\sigma_R{>}0\)) into otherwise collapsed groups, restores group-relative advantages, and anchors updates to a high-quality, low-entropy reference that curbs uncontrolled trajectory growth.

\subsection{Steerable Step-Level Reward Design for Search Tools}
\label{sec:redundancy-reward}

We design our novel \emph{Steerable Step-Level Reward} that alleviates the reward-hacking challenge faced by RLVR training in the multi-turn, tool-interaction setting using vanilla reward (Eq. \ref{eq:vanilla_reward}) as described in (Sec. \ref{Motivation}). Our  reward function enables us to steer (i) \emph{how much} the agent uses tools and (ii) \emph{how} it allocates cognition to exploration and verification. Starting from the vanilla RLVR objective in (Eq. \ref{eq:vanilla_reward}), we make the correctness branch \(R^{\mathrm{answer}}_i\) depend on cognitive behaviors and marginal utility aware labels assigned to each call \(c_t\) in the rollout \(\mathcal{R}=\{(\varphi_t,c_t,o_t)\}_{t=1}^{T}\) by a GPT\mbox{-}4.1 LLM-as-judge as follows:
\[
\begin{array}{@{}l@{}}
\texttt{search\_urls}\in\{\\
\quad \textsc{UniqueSearch:}\;(\text{semantically new query about previously unseen entities/facts}),\\
\quad \textsc{RedundantSearch:}\;(\text{Highly similar to a prior query; overlapping results})\}\\[4pt]
\texttt{query\_url}\in\{\\
\quad \textsc{Exploration:}\;(\text{first query of a new URL}),\\
\quad \textsc{Verification:}\;(\text{cross-source check on a new URL for an existing query; allowed }B_v\text{ times}),\\
\quad \textsc{RedundantQuery:}\;(\text{further checks for a query/fact on new URLs beyond }B_v)\}
\end{array}
\]


From the LLM-as-Judge tool call classification we form tallies\footnote{%
$n_{\text{uniqS}}$ = number of UniqueSearch calls; 
$n_{\text{redS}}$ = number of RedundantSearch calls; 
$n_{\text{explore}}$ = number of Exploration calls; 
$n_{\text{verify}}$ = number of Verification calls; 
$n_{\text{uniqQ}} = n_{\text{explore}} + n_{\text{verify}}$; 
$n_{\text{redQ}}$ = number of RedundantQuery calls.%
} and define the following aggregates:
\begin{equation}
\label{eq:aux}
\rho=\frac{n_{\text{redS}}+n_{\text{redQ}}}{T},\qquad
\Delta_S=n_{\text{uniqS}}-n_{\text{redS}},\qquad
\Delta_Q=n_{\text{uniqQ}}-n_{\text{redQ}}.
\end{equation}

Using these aggregates we define our \emph{Steerable Step-Level Reward} as:
\begin{equation}
\label{eq:redundancy_reward}
r_{i} \;=\;
\begin{cases}
0.1*\,R^{\mathrm{format}}_i \;+\; \textcolor{bad}{\max\!\bigl((1-\rho),\,0.5\bigr)}, & \text{if } a^{(i)}_{\mathrm{pred}} = a_{\mathrm{gt}},\\[4pt]
0.1*\,R^{\mathrm{format}}_i \;+\; \textcolor{good}{c_1*\,\min\!\bigl(1,\tfrac{\Delta_S}{C_S}\bigr)
\;+\; c_2*\,\min\!\bigl(1,\tfrac{\Delta_Q}{C_Q}\bigr)}, & \text{if } a^{(i)}_{\mathrm{pred}} \neq a_{\mathrm{gt}}.
\end{cases}
\end{equation}

 Here, \bad{\(\rho\)} penalizes redundant tool calls \emph{even when the rollout is correct}, pushing for efficiency; whereas \good{\(\Delta_S\)} and \good{\(\Delta_Q\)} provide credit to \emph{incorrect} rollouts that exhibit genuine, non-redundant exploration \& information seeking  behavior.

We set \(c_1=c_2=0.2\), to ensure any incorrect rollout has \(r_i\le 0.5\), while any correct rollout has \(r_i\ge 0.5\), which ensures incorrect trajectories never get rewarded more than the correct ones.  \(c_1{=}c_2\) also ensures equal weight to \texttt{search\_urls} (\(\Delta_S\)) and \texttt{query\_url} (\(\Delta_Q\)).

\paragraph{Steerability.}
We expose three primary knobs:
(i) \(C_S\) and (ii) \(C_Q\) set the saturation thresholds for creditable novelty in \texttt{search\_urls} and \texttt{query\_url}, respectively. Increasing \(C_S\) and/or \(C_Q\) raises the novelty caps, enabling more steps to earn credit when they introduce genuinely new evidence; decreasing them compresses trajectories. (iii) The per-claim verification budget \(B_v\) controls verification depth: higher \(B_v\) permits multiple creditable cross-checks per claim, promoting verification. For our experiments we set \(B_v\) = 1 allowing 1 cross-check per claim, additionally we set \(C_S\) = 8 and \(C_Q\) = 16.

\begin{wrapfigure}{r}{0.45\textwidth}
  \centering
  \vspace{-1 em} 
  \includegraphics[width=\linewidth]{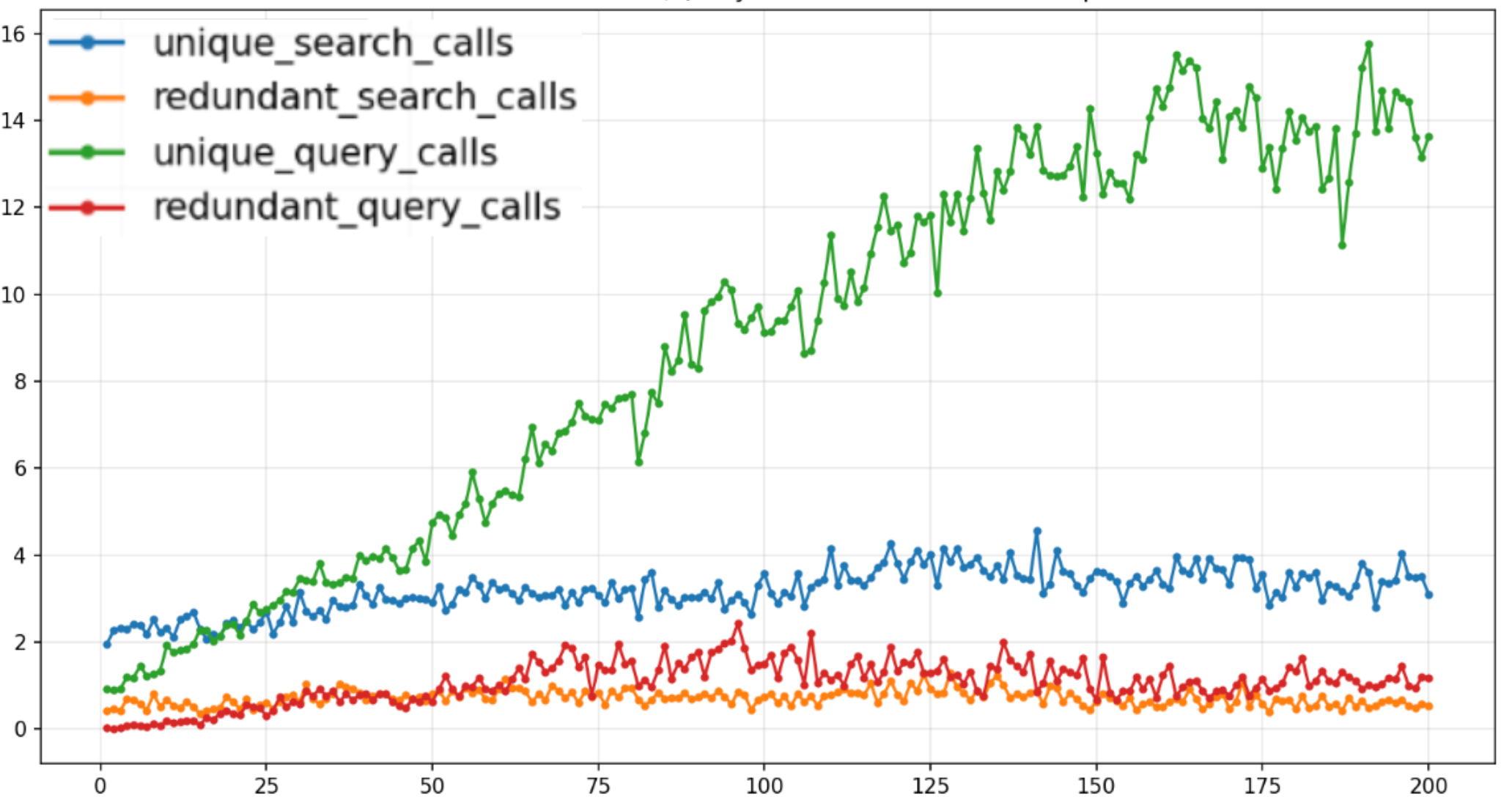}
  \caption{Evolution of unique/redundant tool-calls  during Stage-2 training using our Steerable Step-level Reward (Eq.\ref{eq:redundancy_reward})}
   \vspace{-1 em} 
  \label{fig:sce}
\end{wrapfigure}

\subsection{Training Recipe}

We build our reinforcement learning with verifiable rewards (RLVR) framework on top of \textsc{ReCall}~\citep{recall}. For web search, we use the Serper API~\citep{serperdev}, and implement a retrieval toolchain leveraging Jina-AI together with open-source components such as \textsc{Trafiltura} and \textsc{Crawl4AI}. Training is carried out in two stages.  \textbf{Stage 1.}  
We train with RAPO for 10 epochs on our curated \textsc{DuetQA} dataset, comprising \textbf{4,988} high-quality QA instances. The setup uses a constant learning rate of $1\times 10^{-6}$ with the Adam optimizer ($\beta_1=0.9$, $\beta_2=0.95$), batch size $32$, mini-batch size $16$, $5$ rollouts per group, and top-$p=1.0$ sampling. Each rollout is capped at $32$ tool-interaction steps, with each step limited to $8{,}192$ output tokens. The vanilla reward (Eq. \ref{eq:vanilla_reward}) with $\alpha\ = 0.1$  is used to instill correct tool-calling behavior and strict format adherence.  \textbf{Stage 2.}  
We continue RLVR training for an additional 2 epochs under the same hyperparameter settings. For Stage~2, we construct a mixed dataset by combining \textsc{DuetQA}, with math data from  S1 dataset~\citep{muennighoff2025s1simpletesttimescaling}, and the training split of \textsc{Musique}~\citep{musique}. This combined pool is adversarially filtered against the Stage-1 checkpoint, yielding  \textbf{5,077} instances. From \textsc{Musique}, we retain only questions requiring at least three reasoning hops to ensure sufficient compositional depth. For this stage, we adopt the Steerable Step-Level Reward (Eq. \ref{eq:redundancy_reward}) to extend the tool-use horizon beyond $20$ calls in a stable manner.  We use the Qwen3-4B model~\citep{yang2025qwen3technicalreport} as the base, which supports a maximum context length of $40{,}960$ tokens; we utilize the full window during training. We use GPT-4.1-mini(Temperature=0.) as the query LLM for training and evaluation unless stated otherwsie.
A higher sampling temperature of $1.4$ is applied to Qwen3 models, consistent with prior findings~\citep{Polaris2025}. All experiments are conducted on a single node with $8\times$H100 GPUs.

\begin{table}[t]
  \centering
  \setlength{\tabcolsep}{3.8pt}
  \caption{Accuracy(\%) of \emph{Fathom-Search-4B} on DeepSearch benchmarks SimpleQA, FRAMES, WebWalker, Seal0, Musique and general reasoning benchmarks HLE, AIME-25, GPQA-D, MedQA. ‘Avg’ is the unweighted mean within each block. Bold/italics denote best/second-best per benchmark.}

  \label{tab:final-eval-1}
  \resizebox{\textwidth}{!}{
  \begin{tabular}{lcccccccccccc}
    \toprule
    & \multicolumn{6}{c}{\textbf{DeepSearch Benchmarks}} & \multicolumn{5}{c}{\textbf{General Reasoning Benchmarks}} \\
    \cmidrule(lr){2-7} \cmidrule(lr){8-12}
    \textbf{Model} & \textbf{SimpleQA} & \textbf{FRAMES} & \textbf{WebWalker} & \textbf{Seal0} & \textbf{Musique} & \textbf{Avg} 
                  & \textbf{HLE} & \textbf{AIME-25} & \textbf{GPQA-D} & \textbf{MedQA} & \textbf{Avg} \\
    \midrule
    \multicolumn{12}{c}{\textbf{Closed-Source Models}}\\
    \midrule
    \addlinespace[2pt]
    GPT-4o (without search)  & 34.7 & 52.4 & 3.2  & 7.2  & 34.0 & 26.3 & 2.3 & \emph{71.0} & \emph{53.0} & \emph{88.2} & 53.6 \\
    o3 (without search)      & 49.4 & 43.2 & 14.0 & 14.0 & \emph{48.9} & 33.9 & \emph{20.3} & \textbf{88.9} & \emph{85.4} & \textbf{95.4} & \emph{72.5} \\
    GPT-4o (with search)     & \emph{84.4}  & \emph{63.7} & \emph{31.6} & \emph{15.3} & 37.5 & \emph{46.5} & 4.3 & \emph{71.0} & \emph{53.0}  & \emph{88.2} & 54.1 \\
    o3 (with search)         & \textbf{96.0} & \textbf{86.8} & \textbf{57.0} & \textbf{49.5} & \textbf{51.2} & \textbf{68.1} & \textbf{27.4} & \textbf{88.9} & \textbf{85.4} & \textbf{95.4} & \textbf{74.3} \\
    \midrule
    \multicolumn{12}{c}{\textbf{Open-Source Models}}\\
    \midrule
    \addlinespace[2pt]
    Qwen-2.5-7B          & 3.96 & 16.5 & 2.1 & 1.4 & 6.2 & 6.0 & 1.2 & 10 & 33.0 & 61.2 & 24.7 \\
    Qwen-2.5-7B + Search & 50.8 & 23.3 & 10.1 & 3.0 & 13.6 & 20.2 & 2.4 & 10 & 33.5 & 62.0 & 25.3 \\
    Qwen3-4B             & 3.8  & 14.7 & 2.6 & 2.1 & 9.0 & 6.4 & 4.2 & \emph{65.0} & 55.1 & 71.0 & 48.8 \\
    Qwen3-4B + Search    & 67.7 & 27.2 & 17.5 & 6.2 & 18.7 & 27.5 & 6.2 & \emph{65.0} & \emph{55.9} & 72.0 & 49.8 \\
    ZeroSearch-3B        & 51.9 & 11.3 & 8.7  & 7.1 & 13.8 & 18.6 & 3.4 & 10.0 & 14.6 & 51.0 & 17.3 \\
    ZeroSearch-7B        & 75.3 & 30.0 & 18.2 & 6.2 & 20.6 & 30.1 & 4.2 & 10.0 & 29.3 & 57.5 & 22.8 \\
    R1-Searcher-7B         & 58.8 & 37.0 & 1.8  & 1.4 & 19.1 & 23.6 & 2.1 & 10.0 & 33.3 & 56.5 & 25.5 \\
    search-o1 (Qwen3-4B) & 57.5 & 26.8 & 10.8 & 5.5 & 15.3 & 23.2 & 3.4 & 40.0 & 30.5 & 53.7 & 31.9 \\
    WebSailor-3B         & 87.1 & 44.4 & \textbf{52.2} & 9.0 & 27.4 & 44.0 & 7.4 & 40.0 & 45.5 & 51.3 & 36.0 \\
    Jan-Nano-32K         & 80.7 & 36.1 & 25.0 & 6.2 & 21.4 & 33.9 & 5.5 & 60.0 & 37.4 & 66.0 & 42.2 \\
    Jan-Nano-128K        & 83.2 & 43.4 & 33.7 & 6.2 & 23.9 & 38.1 & 6.1 & 53.3 & 51.0 & 65.4 & 44.0 \\
    II-Search-4B         & \emph{88.2} & \emph{58.7} & 40.8 & 17.1 & \emph{31.8} & \emph{47.3} & \emph{7.4} & 60.0 & \emph{51.5} & \emph{72.1} & 47.8 \\
    \midrule
    \rowcolor{blue!15}  Fathom-Search-4B (Stage-1) & 88.1 & 57.2 & 39.0 & \emph{19.8}& 31.3 & 47.1 & 6.7 & 60.0 & 55.6 &\textbf{75.4} & \emph{49.4} \\
    \rowcolor{blue!20} Fathom-Search-4B (Stage-2)& \textbf{90.0} & \textbf{64.8} & \textit{50.0} & \textbf{22.5} & \textbf{33.2} & \textbf{52.1} & \textbf{9.5} & \textbf{70.0} & \textbf{60.1} & \textbf{75.4} & \textbf{53.8} \\
    \bottomrule
  \end{tabular}
  }
\end{table}

\section{Fathom-Synthesizer-4B}
\textit{Fathom-Synthesizer-4B} is a planning and synthesis model built on Qwen3-4B via supervised fine-tuning (SFT). It converts multi-hop DeepSearch traces from \emph{Fathom-Search-4B} into decision-grade, citation-dense \emph{DeepResearch Reports}. Following a \emph{Plan-then-Write} protocol, the model first decomposes the question into sub-goals, defines the report structure, maps evidence to sections, and specifies strategies for insight generation; only then does it produce the public report with citations drawn strictly from URLs explored by \emph{Fathom-Search-4B}. This explicit planning improves question alignment, strengthens citation accuracy through section-level constraints, and provides structured supervision during SFT, enhancing the distillation process.

\subsection{DeepResearch-SFT}
\textsc{DeepResearch-SFT} is a synthetic dataset distilled from GPT-5 \citep{o3_o4mini} to train \textit{Fathom-Synthesizer-4B}, it provides  supervision along three complementary axes:
\textbf{(i) Question decomposition.}
Each input question $q$ is decomposed into ordered sub-questions $\pi^{\text{decomp}}=(S_1,\ldots,S_n)$, which form the report scaffold and ensure coverage of all facets\textbf{(ii) Section mapping.}
Every piece of evidence recovered during search (URLs, quoted passages, tables, figures) is grounded to one or more sections via a mapping $\pi^{\text{map}}$, this aligns each explored URL to the most relevant $S_i$, enhancing citation accuracy and preventing omissions/duplication.\textbf{(iii) Planning for insights.}
The model specifies an analysis strategy $\pi^{\text{insight}}$ how the gathered evidence should be synthesized into higher-level insights. Formally, given a question \(q\) and trajectory \(\tau=\{\mathcal{R}_1,\ldots,\mathcal{R}_T\}\), the teacher outputs \texttt{Plan} and \texttt{Report}. The plan \(\pi=(\pi^{\text{decomp}},\pi^{\text{map}},\pi^{\text{insight}})\) appears in a private \texttt{<think>} block, followed by the public report \(r\). The training target is \(y=\texttt{<think>}\;\pi\;\texttt{</think>}\;r\). \noindent\textbf{Report structure.}
The public-facing report $r$ follows a fixed, inline-citation–driven format: an \emph{Executive Summary} followed by a \emph{Main Body} organized exactly by the sections $(S_1,\ldots,S_n)$ from $\pi^{\text{decomp}}$, where each section weaves the mapped evidence from $\pi^{\text{map}}$ using the analysis strategy in $\pi^{\text{insight}}$. Sections are citation-dense: every pivotal or non-obvious claim carries inline citations drawn \emph{only} from URLs explored in $\tau$, with section-level citations restricted to items mapped to that section in $\pi^{\text{map}}$. The report concludes with a deduplicated \emph{Sources used} list of the cited URLs.
\textbf{Thematic diversity \& scale.}
Training questions are generated via the \emph{Seeded Question mode} (Sec~\ref{sec:dataset-curation}), starting from 100 open-ended real-world questions spanning law, business, technology, science, and policy. Rewritten across sampled themes, this yields \textbf{2{,}500} questions for training.

\subsection{Training Recipe}
We fine-tune \textbf{Qwen3-4B} on \textsc{DeepResearch-SFT}, training for \textbf{5 epochs} on the 2{,}500-sample split using a single node of $8\times$H100 GPUs. We use \texttt{bf16}, FlashAttention-2, a $65{,}536$-token context, gradient accumulation of $8$, cosine LR with peak $5.0\times 10^{-5}$, Adam ($\beta_1{=}0.9,\beta_2{=}0.95$), and sequence parallel size $4$.\textbf{Context extension.} Our DeepSearch traces exhaust Qwen3-4B’s native $40{,}960$-token context window, so we \emph{extend} the effective context during SFT using YaRN RoPE scaling: \texttt{rope\_scaling:\{type=yarn, factor=2.0\}}. This increases the usable positional range to $65{,}536$ tokens, allowing the synthesizer to ingest the full investigation trace and generate high-quality synthesis while preserving section alignment and citation locality.
\label{sec:results}

\begin{table}[t]
  \centering
  \scriptsize
  \setlength{\tabcolsep}{3.8pt}
  \caption{\textbf{Accuracy(\%) of various Open/Closed-sourced DeepResearch-Agents and Search Augmented LLMs on DeepResearch-Bench.} Bold/italics denote best/second-best per category.}
  \label{tab:deepresearch-overall}
  \resizebox{\textwidth}{!}{
  \begin{tabular}{lrrrrrrr}
    \toprule
     &  \multicolumn{5}{c}{\textbf{RACE}} & \multicolumn{2}{c}{\textbf{FACT}} \\
    \cmidrule(lr){2-6} \cmidrule(lr){7-8}
     \textbf{Model} & \textbf{Overall} & \textbf{Comp.} & \textbf{Depth} & \textbf{Inst.} & \textbf{Read.} & \textbf{C. Acc.} & \textbf{E. Cit.} \\
    \midrule
    \multicolumn{8}{c}{\textbf{Closed Source LLM with Search Tools}} \\
    \midrule
    Claude-3.7-Sonnet w/Search            & \textbf{40.67} & \textbf{38.99} & \textbf{37.66} & \textbf{45.77} & 41.46 & \textit{93.68} & \textit{32.48} \\
    Perplexity-Sonar-Reasoning-Pro (high) & \textit{40.22} & \textit{37.38} & 36.11 & \textit{45.66} & \textbf{44.74} & 39.36 & 8.35 \\
    Gemini-2.5-Pro-Grounding              & 35.12 & 34.06 & 29.79 & 41.67 & 37.16 & 81.81 & \textbf{32.88} \\
    GPT-4o-Search-Preview (high)          & 35.10 & 31.99 & 27.57 & 43.17 & 41.23 & 88.41 & 4.79 \\
    GPT-4.1 w/Search (high)               & 33.46 & 29.42 & 25.38 & 42.33 & 40.77 & 87.83 & 4.42 \\
    
    \midrule
    \multicolumn{8}{c}{\textbf{Closed Source Deep Research Agent}} \\
    \midrule
    Grok Deeper Search                    & 40.24 & 37.97 & 35.37 & 46.30 & 44.05 & \textit{83.59} & 8.15 \\
    Perplexity-DeepResearch              & 42.25 & 40.69 & 39.39 & 46.40 & 44.28 & \textbf{90.24} & 31.26 \\
    Gemini-2.5-Pro DeepResearch          & \textbf{48.88} & \textbf{48.53} & \textbf{48.50} & \textit{49.18} & \textbf{49.44} & 81.44 & \textbf{111.21} \\
    OpenAI-DeepResearch                  & \textit{46.98} & \textit{46.87} & \textit{45.25} & \textbf{49.27} & \textit{47.14} & 77.96 & \textit{40.79} \\    
    \midrule
    \multicolumn{8}{c}{\textbf{Open Source Deep Research Agent}} \\
    \midrule
     Kimi-Researcher  & \textit{44.64} & \textbf{44.96} & \textit{41.97} & 47.14 & \textit{45.59} & -- & -- \\
    Doubao-DeepResearch & 44.34 & 44.84 & 40.56 & 47.95 & 44.69 & \textit{52.86} & \textbf{52.62} \\
    LangChain Open-DeepResearch & 43.44 & 42.97 & 39.17 & \textit{48.09} & 45.22 & -- & -- \\
    
    \rowcolor{blue!20} Fathom-DeepResearch  & \textbf{45.47} & \textit{42.98} & \textbf{45.14} &\textbf{ 48.25} & \textbf{46.12} & \textbf{56.1} & \textit{38.3} \\
    \bottomrule

  \end{tabular}
  }
\end{table}

\section{Baselines, Benchmarks \& Metrics}
\label{sec:baselines}

\textbf{Baselines.} \textbf{Open-source} DeepSearch agents: with public checkpoints: Jan-Nano~\citep{dao2025jannanotechnicalreport}, II-Search-4B ~\citep{II-Search-4B}, Qwen3-4B ~\citep{yang2025qwen3technicalreport}, ZeroSearch~\citep{sun2025zerosearchincentivizesearchcapability}, Search-o1~\citep{searcho1}, R1-Searcher~\citep{song2025r1searcherincentivizingsearchcapability}, WebSailor~\citep{li2025websailornavigatingsuperhumanreasoning}. \textbf{Closed-source:} comparators: o3~\citep{o3_o4mini}, GPT-4o~\citep{gpt4o}.

\textbf{Benchmarks (9).} \textbf{DeepSearch (5):} SimpleQA~\citep{simpleqa}, FRAMES~\citep{frames}, WebWalkerQA~\citep{wu2025webwalkerbenchmarkingllmsweb}, Seal0~\citep{sealqa}, MuSiQue~\citep{musique}. 
\textbf{General reasoning (4):} HLE~\citep{phan2025humanitysexam}, AIME-25~\citep{aime25}, GPQA-Diamond~\citep{rein2024gpqa}, MedQA~\citep{jin2021medqa}. 
\textit{Metric:} Pass@1 using GPT-4.1-mini LLM as Judge (Temperature=0). 
\textbf{DeepResearch (1):} DeepResearch-Bench~\citep{deepresearchbench}. 
\emph{Metrics:} RACE (reference-based adaptive criteria-driven evaluation of comprehensiveness, depth, instruction-following, readability) and FACT (factuality via citation accuracy and effective citation count) for open-ended citation driven report generation.

\begin{figure*}[]
  \centering
  \includegraphics[width=\linewidth]{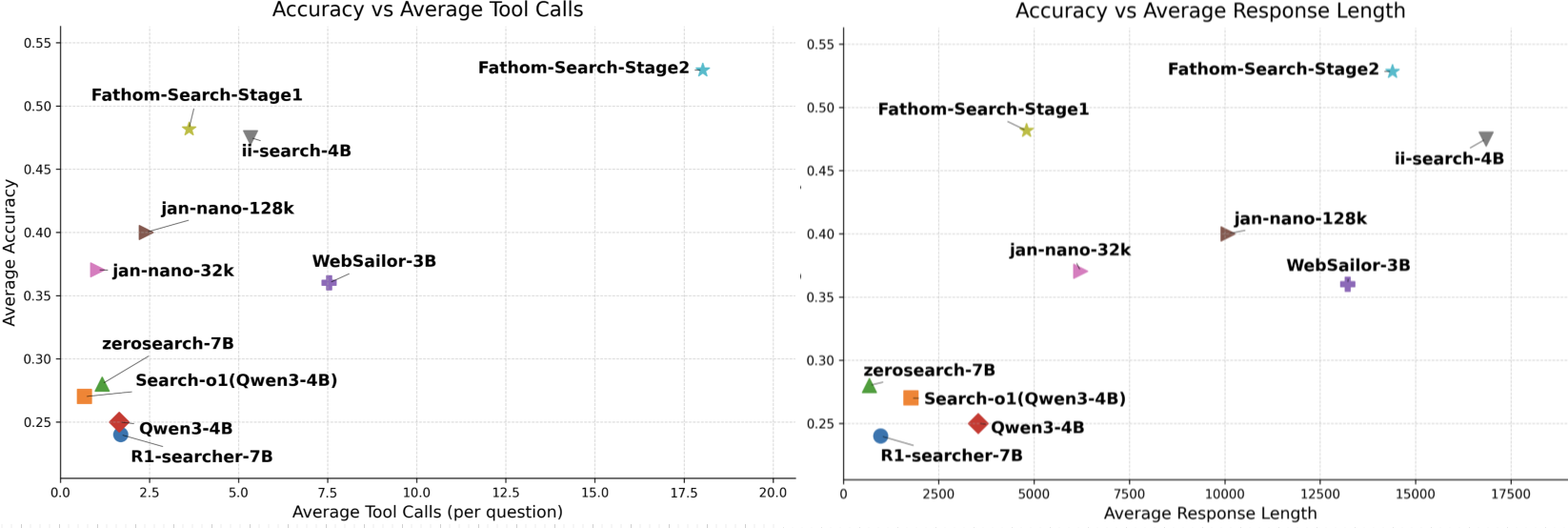}
\caption{\textbf{Accuracy vs Response length} \& \textbf{Accuracy vs  Avg.Tool calls} plot comparing open-source DeepSearch models, clearly demonstrates the higher accuracy and efficient long horizon tool interaction ability of Fathom-Search models compared to its contemporaries. (\emph{Note:} Here accuracy refers to the unweighted average of all benchmarks in Table. \ref{tab:final-eval-1}.)}
\label{fig:fml2}
\vspace{-1em}
\end{figure*}

\section{Discussion}

\begin{table*}[t]
  \centering
  \scriptsize
   \setlength{\tabcolsep}{8 pt}
  \caption{\textbf{Ablation} on using   RAPO as the policy optimization algorithm for Stage-1 training compared to GRPO. Both trainings done on top of Qwen3-4B model.}
  \label{tab:rapo-grpo}
  \begin{tabular}{lccccc}
    \toprule
    \textbf{Algorithm} & \textbf{SimpleQA} & \textbf{FRAMES} & \textbf{WebWalker} & \textbf{Seal0} & \textbf{Avg. Tokens} \\
    \midrule
    \textbf{GRPO}   & 87.8 & 55.2 & 33.8 & 14.4 & 9,000 \\
    \textbf{RAPO}   & 88.1  & 57.2 & 39.0  & 19.8  & 5,000 \\
    \bottomrule
  \end{tabular}
\end{table*}

\begin{table*}[t]
  \centering
  \scriptsize
  \setlength{\tabcolsep}{8 pt}
  \caption{\textbf{Ablation} on using our steerable step-level reward compared to the vanilla trajectory-level RLVR reward for Stage-2 training. Trained on top of Fathom-Search-4B (Stage-1)  using RAPO.}
  \label{tab:steerable-reward-ablation}
  \begin{tabular}{lccccc}
    \toprule
    \textbf{Reward} & \textbf{SimpleQA} & \textbf{FRAMES} & \textbf{WebWalker} & \textbf{Seal0} & \textbf{Avg. Tokens} \\
    \midrule
    \textbf{Vanilla Reward (Eq. ~\ref{eq:vanilla_reward}}) & 88.2 & 58.2 & 43.2 & 21.6  & 5,500\\
    \textbf{Steerable Step-Level Reward (Eq. ~\ref{eq:redundancy_reward}})     & 90 & 64.8 & 50 & 22.5 & 14,500 \\
    \bottomrule
  \end{tabular}
\end{table*}


\paragraph{Strong performance rivaling closed-source proprietary models} 
Fathom-DeepResearch establishes itself as a clear state-of-the-art by achieving large, non-incremental gains on the most challenging DeepSearch tasks like \textit{FRAMES, WebWalker, \& Seal0}, (Table. \ref{tab:final-eval-1}), while also showing strong generalization to broader reasoning benchmarks like (GPQA-Diamond and Humanity's Last Exam). Unlike many search-augmented systems that falter outside their training domain, it consistently outperforms both its base model and other open-source systems, and even surpasses larger closed-source models such as GPT-4o with notable margins. On open-ended benchmark: \textit{DeepResearch-Bench}, it outperforms most proprietary closed-source systems (including Claude, Grok, and Perplexity Deep Research) (Table. \ref{tab:deepresearch-overall}) underscoring its competitiveness in end-to-end deep research tasks.

\begin{wrapfigure}{r}{0.45\textwidth}
  \vspace{-1.5em}
  \includegraphics[width=\linewidth]{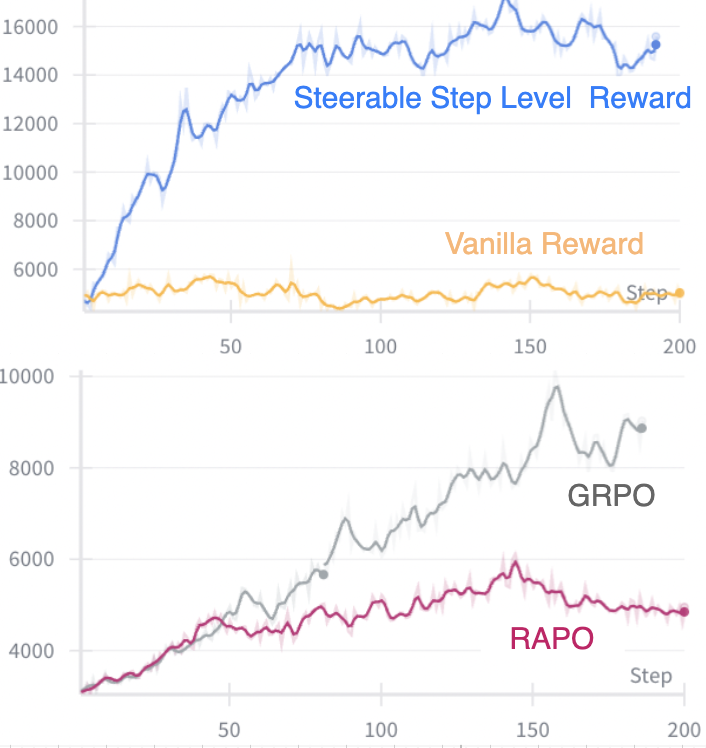}
  \caption{Response length evolution during 
   \textbf{(i) top.} Stage-2 training (Steerable Step Level Reward vs. Vanilla Reward) \& \textbf{(ii) bottom.} Stage-1 training (GRPO vs RAPO)}
   \vspace{-3em}
  \label{fig:tts}
\end{wrapfigure}

\paragraph{On policy optimization: RAPO vs.\ GRPO.}  
Table~\ref{tab:rapo-grpo} contrasts RAPO and GRPO as the policy-optimization algorithm for Stage-1 training. With the Stage-1 setup fixed, RAPO consistently outperforms GRPO across DeepSearch benchmarks, This shows that RAPO provides a more stable and effective optimization signal. As shown in Fig.~\ref{fig:tts}, GRPO expands response length as training progresses, but this growth does not translate into higher accuracy because of the model collapsing into redundant tool-spamming behavior. RAPO, in contrast, achieves stronger efficiency and more accurate results.

\textbf{On the Steerable Step-Level reward.}
As shown in Fig.~\ref{fig:sce}, the steerable step-level reward provides a finer-grained training signal that steers the tool calling behavior of the model. By directly shaping the utility of each intermediate step, it encourages controlled growth in response length without inflating reasoning traces with redundant tool call spam, thereby yielding both efficiency and stability in multi-step reasoning, outperforming the vanilla RLVR reward function on all DeepSearch tasks as shown in Table. \ref{tab:steerable-reward-ablation}

\textbf{Limitations.}
While RAPO is effective for stabilizing multi-turn RL training, it shows limited test-time scaling. As illustrated in Fig.~\ref{fig:tts}, RAPO with vanilla rewards during Stage-2 training saturates before 6,000 tokens and yields only marginal accuracy gains (Table~\ref{tab:steerable-reward-ablation}) as question difficulty increases, when the steerable step-level reward is absent. This trade-off arises from its reliance on trajectory replacement in the replay buffer, which anchors learning to low-entropy traces which prevents training collapse but also hinders adaptation to extended reasoning horizons. More broadly, our current system depends on synchronous training pipelines that, although simple to implement, remain inefficient and brittle at scale. Transitioning to asynchronous frameworks presents a natural next step for improving efficiency and robustness.

\section{Conclusion}
We present Fathom-DeepResearch, an agentic system that addresses critical gaps in open-source deep research capabilities through two specialized 4B models: Fathom-Search-4B for multi-turn web search and reasoning, and Fathom-Synthesizer-4B for structured report synthesis. Our key contributions include DuetQA, a multi-agent self-play dataset that ensures search dependency; RAPO, a stabilized extension of GRPO that enables reliable tool use beyond 20 calls through curriculum pruning, advantage scaling, and replay buffers; a steerable step-level reward system that mitigates reward hacking while providing explicit control over exploration and verification behaviors; and DeepResearch-SFT, a synthetic corpus that enables comprehensive information synthesis through explicit plan-then-write supervision.

\bibliography{iclr2025_conference}
\bibliographystyle{iclr2025_conference}

\appendix

\end{document}